\documentclass[graybox]{svmult}

\usepackage{mathptmx}       
\usepackage{helvet}         
\usepackage{courier}        
\usepackage{type1cm}        
\usepackage{makeidx}         
\usepackage{graphicx}        
\usepackage{multicol}        
\usepackage[bottom]{footmisc}
\usepackage{multirow}

\makeindex             

\usepackage{algorithm2e}
\usepackage{xcolor}
\usepackage{easyeqn}
\usepackage{bm}

\def\T{\mathrm{T}}
\def\nsize{n}
\def\adjacencyMatrix{\mathbf{A}}
\def\adjacencyEigenvectors{\mathbf{U}}
\def\adjacencyEigenvalues{\mathbf{L}}
\def\simDiagMatrix{\boldsymbol{\Lambda}}
\def\lambdav{\boldsymbol{\lambda}}

\begin{document}

\title*{Simultaneous Matrix Diagonalization for Structural Brain Networks Classification}

\author{Nikita Mokrov, Maxim Panov, Boris A. Gutman, Joshua I. Faskowitz, Neda Jahanshad and Paul M. Thompson}
\institute{Nikita Mokrov \at Moscow Institute of Physics and Technology, Institute for Information Transmission Problems of RAS, Moscow, Russia, \email{mokrov.ns@phystech.edu}
\and
Maxim Panov \at Skolkovo Institute of Science and Technology (Skoltech), Institute for Information Transmission Problems of RAS, Moscow, Russia, \email{m.panov@skoltech.ru}
\and
Boris A. Gutman, Joshua I. Faskowitz, Neda Jahanshad, Paul M. Thompson \at Imaging Genetics Center, Stevens Neuroimaging and Informatics Institute, University of Southern California, Marina del Rey, USA
}

\maketitle

\abstract{
  This paper considers the problem of brain disease classification based on connectome data. A connectome is a network representation of a human brain. The typical connectome classification problem is very challenging because of the small sample size and high dimensionality of the data. We propose to use simultaneous approximate diagonalization of adjacency matrices in order to compute their eigenstructures in more stable way. The obtained approximate eigenvalues are further used as features for classification. The proposed approach is demonstrated to be efficient for detection of Alzheimer's disease, outperforming simple baselines and competing with state-of-the-art approaches to brain disease classification.
}

\section{Introduction}
\label{sec: introduction}
  Connectomics is becoming one of the driving directions of modern neuroscience research. The term ``connectome'' was introduced in 2005 to describe network representations of human brain~\cite{Sporns2005,Hagmann2005}. Formally, a connectome is a weighted graph where each node represents certain part of a brain and each edge characterizes the structural connection between the regions of a brain. Insights about the network organization of the brain can help to detect disease-related changes in their early stages. 

  The main challenge of connectome classification is a combination of small sample size with high dimensionality of the data. Especially challenging is the fact that each object in the dataset is represented by graph. That is why most of the methods for feature extraction from connectome graphs process each network separately. In this work, we propose different approach to obtain more adequate and stable features based on the joint structure of all connectomes in the dataset. We employ simultaneous diagonalization of adjacency matrices to compute and use values on the diagonal as features for classification.

  The rest of the paper is organized as follows. Section~\ref{sec: classification} discusses the statement of considered classification problem and existing approaches to its solution. Section~\ref{sec: proposedMethod} introduces the proposed feature extraction approach as well as overall classification pipeline. In Section~\ref{sec: experiments} we describe the experimental setup and the results of experiments. Section~\ref{sec: conclusions} summarizes the study and discusses possible future research directions.

\section{Brain networks classification}
\label{sec: classification}
\subsection{Problem statement}
\label{sec: problemStatement}
  We consider the classification problem where a single data point is described by the adjacency matrix of the connectome network. The dataset becomes the set of \(\nsize\) pairs
  \begin{EQA}[c]
    D = \bigl\{(\adjacencyMatrix_1, y_1), \dots, (\adjacencyMatrix_{\nsize}, y_{\nsize})\bigr\},
  \end{EQA}
  where \(\adjacencyMatrix_i\) is the adjacency matrix of \(i\)-th connectome and \(y_i\) is the corresponding class label. 
  We are going to consider several binary classification problems with \(y_i \in \{0, 1\}\).
  The goal of the study is to construct the classification algorithm which takes adjacency matrix \(\adjacencyMatrix\) of a connectome as input and outputs predicted class label \(\hat{y}\). 

\subsection{Existing approaches}
\label{sec: existingApproaches}
  The problem of brain network classification has been paid much attention recently~\cite{Kurmukov2016,Dodonova2016,Petrov2016,Kurmukov2016b,Kurmukov2017}. This problem is non-trivial as most modern classification algorithms can work only with vectorial data while in our case each object in the dataset is represented by graph. Thus, one needs to generate numerical features from a graph, which can be done in multiple ways.

  The first important question is what representation of a graph to use for further processing. The paper~\cite{Petrov2016} considers different types of matrices such as adjacency matrix, graph Laplacian, normalized graph Laplacian and others. They further consider different types of processing for the chosen matrix:
  \begin{itemize}
    \item binarization;
    \item geometric normalization by the distance between centers of brain regions;
    \item weighted communicability normalization~\cite{Crofts2009}.
  \end{itemize}
  Each of the choices above has certain pros and cons, however the authors of~\cite{Petrov2016} report that the best results are achieved by combination of geometric and weighted communicability normalization.

  The second important question is what to do with the matrix in order to use it as an input to a classification algorithm. Several approaches have been considered in the literature, including vectorization of the whole matrix, considering the vector of node degrees of a graph~\cite{Petrov2016} or the vector of eigenvalues~\cite{Dodonova2016}, and different variants of constructing the kernel between graphs~\cite{Kurmukov2016,Dodonova2016}. We note, that in all these approaches features are generated for each connectome independently. To the best of our knowledge, the only approach proposed for collective feature extraction in brain networks is higher-order SVD~\cite{Zhan2015}.

  If we consider the problem of feature extraction from general prospective, we almost unavoidably come to the question of finding structure in the data. The key challenge in the majority of machine learning problems is dealing with high dimensionality of the data, given a sample of very limited size. In general case, so-called ``curse of dimensionality'' requires the sample size to be of super linear order compared to the dimension of the problem. However, in many situations the exploitation of hidden structure in data enables us to significantly decrease the number of samples needed to achieve a desired level of quality. For the matrix data, there are several types of useful structure, in particular
  \begin{itemize}
    \item block structure~\cite{Rudie2013,Kurmukov2016,Kurmukov2016b};
    \item low-rank structure~\cite{Rohde2011}.
  \end{itemize}
  In the graph case, blocks correspond to the community structure. By community structure we mean the existence of groups of nodes that are much more strongly linked inside the group compared to the rest of the graph. We note that the community detection approach has been recently used for connectome classification~\cite{Kurmukov2016b,Kurmukov2017}, where a certain kernel function was used to measure similarity between graph partitions, which were again obtained independently for each graph.

  In this paper, we stick to simply processing the adjacency matrix and focus more on the feature extraction step. The main contribution of the study is dealing with joint structure of matrices in the dataset as opposed to usual exploitation of individual properties. We consider an approach to feature extraction based on spectral properties of set of adjacency matrices which is conceptually simple, but allows to achieve state-of-the-art results.

\section{Proposed method}
\label{sec: proposedMethod}
  The proposed classification pipeline consists of two steps.
  \begin{enumerate}
    \item Feature extraction by simultaneous matrix diagonalization.
    \item Classification based on obtained features.
  \end{enumerate}
  We are going to discuss these steps below.

\subsection{Simultaneous matrix diagonalization}
\label{sec: matrixDiagonalization}

  We start by recalling the classical eigenvalue decomposition of the adjacency matrix:
  \begin{EQA}[c]
  \label{eq: SVD}
     \adjacencyMatrix_i = \adjacencyEigenvectors_i \adjacencyEigenvalues_i \adjacencyEigenvectors_i^{\T}, ~ i = 1, \dots, \nsize,
  \end{EQA} 
  where \(\adjacencyEigenvectors_i\) is a matrix of eigenvectors of the matrix~\(\adjacencyMatrix_i\), and \(\adjacencyEigenvalues_i\) is a diagonal matrix of corresponding eigenvalues.

  Both eigenvectors \(\adjacencyEigenvectors_i\) and eigenvalues \(\adjacencyEigenvalues_i\) can be useful for classification. For example, eigenvectors \(\adjacencyEigenvectors_i\) can be used to construct a kernel on linear subspaces corresponding to the top eigenvalues~\cite{Hamm2008}, while eigenvalues \(\adjacencyEigenvalues_i\) can be used directly as feature vectors for machine learning algorithms. The latter approach was successfully used for connectome classification~\cite{Dodonova2016}. However, it was found in~\cite{Dodonova2016} that eigenvalues themselves are very noisy and certain smoothing is needed to achieve reasonable results.

  In our approach, we are going to look for linear subspaces that are common for all the matrices in the dataset. The goal is achieved by simultaneous matrix diagonalization, i.e. the following decomposition
  \begin{EQA}[c]
  \label{eq: sim_diag}
     \adjacencyMatrix_i = \adjacencyEigenvectors \simDiagMatrix_i \adjacencyEigenvectors^{\T}, ~ i = 1, \dots, \nsize,
  \end{EQA}
  where \(\adjacencyEigenvectors\) is an orthogonal matrix joint for all the adjacency matrices \(\adjacencyMatrix_i\), while approximately diagonal matrices \(\simDiagMatrix_i\) are individual.

  We note that a set of matrices is simultaneously diagonalizable if and only if all the matrices in the set commute. Unfortunately, this property doesn't hold for the typical adjacency matrices of connectome networks. That is why only approximate simultaneous diagonalization of adjacency matrices is possible. There are number of algorithms designed specially for this problem~\cite{Yeredor2002,Cardoso1996,Tichavsky2009}. We are going to use the method~\cite{Cardoso1996}, which minimizes the sum of off-diagonal elements
  \begin{EQA}[c]
    F(\adjacencyEigenvectors) = \sum_{i = 1}^{\nsize} \mathrm{off}(\adjacencyEigenvectors^{-1} \adjacencyMatrix_i \adjacencyEigenvectors^{-\T}), ~~ \mathrm{off}(\mathbf{M}) = \sum_{k \neq l} m_{kl}^2
  \end{EQA}
  over \(\adjacencyEigenvectors\) via plane rotations. We further use diagonals of obtained matrices
  \begin{EQA}[c]
    \lambdav_i = \mathrm{diag}(\simDiagMatrix_i) = \mathrm{diag}(\adjacencyEigenvectors^{-1} \adjacencyMatrix_i \adjacencyEigenvectors^{-\T})
  \end{EQA}
  as features for classification.

\subsection{Choice of classifier}
  Having selected the approach of feature extraction from connectome networks, one question remaining is the choice of particular algorithm to perform classification based on the obtained features. We consider several classical algorithms such as Logistic Regression (LR)~\cite{Hosmer2013}, Support Vector Machines (SVM) with linear kernel~\cite{Cortes1995}, Linear Models with Elastic Net Regularization trained by stochastic gradient descent (SGD)~\cite{Zou2005}, Random Forest (RF)~\cite{Breiman2001} and Gradient Boosted Decision Trees (GBDT)~\cite{Friedman2001}.

\section{Experiments}
\label{sec: experiments}
\subsection{Data}
\label{sec: data}
  In this work, we consider the data gathered in the framework of the Alzheimer's Disease Neuroimaging Initiative. The database ADNI2 consists of MRI scans of 228 individuals with 756 scans in total. The mean age of the people in entire cohort was \(72.9 \pm 7.4\) years (132 men and 96 women). This cohort includes people with Alzheimer's disease (AD), early-stage and late-stage mild cognitive impairment (EMCI and LMCI) and normal controls (NC). The sizes of the groups are summarized in Table~\ref{tab: groupSizes}.

  \begin{table}
    \centering
    \label{tab: groupSizes}
    \begin{tabular}{|c|c|c|c|c|}
      \hline
      Number & AD & LMCI & EMCI & NC \\
      \hline
      Individuals & $47$ & $40$ & $80$ & $61$ \\
      \hline
      Scans & $136$ & $147$ & $283$ & $190$ \\
      \hline
    \end{tabular}
    \caption{Subdivision of ADNI2 data samples into 4 categories.}
  \end{table}

  T1-weighted (T1w) images were processed with FreeSurfer~\cite{Fischl2012}, where we used cortical parcellation based on the Desikan-Killiany atlas~\cite{Desikan2006} of 68 cortical brain regions. In parallel, the average \(b_0\) of the DWI images was registered to the downsampled (2mm isotropic MNI) T1w image, to account for susceptibility artifacts. DWI images were corrected for eddy current and motion related distortions; b-vectors were rotated accordingly. Probabilistic streamline tractography was performed using the Dipy~\cite{Garyfallidis2014} LocalTracking module with constrained spherical deconvolution (CSD)~\cite{Tax2014}. Streamlines longer than 5mm with both ends intersecting the cortical surface were retained. Edge weights in the original cortical connectivity matrices were thus proportional to the number of streamlines detected by the algorithm.

  We also tried the proposed approach on~\textit{UCLA Autism dataset}~\cite{Rudie2013} and~\textit{UCLA APOE-4 dataset}~\cite{Brown2012,Brown2011}. We note that for the first dataset there are 94 connectivity matrices with 264 brain regions while the second dataset contains 55 individuals with brain partitioned in 110 regions. Our diagonalization method failed to simultaneously diagonalize matrices for both datasets. The possible explanation is in much higher dimensionality combined with greater noise in the data. The latter statement is partially supported by the fact that typical quality of classification for both datasets~\cite{Petrov2016,Dodonova2016} is much lower than for ADNI2 database considered in this work.

\subsection{Setup of experiments}
\label{sec: setup}
  We consider four binary classification problems:
  \begin{enumerate}
    \item Alzheimer's disease versus normal controls (AD vs NC);
    \item Alzheimer's disease versus late-stage mild impairment (AD vs LMCI);
    \item late-stage mild impairment vs early-stage mild impairment  (LMCI vs EMCI);
    \item early-stage mild impairment vs normal contact  (EMCI vs NC).
  \end{enumerate}
  We note that the first problem is the most simple one, while the other three are harder but more important in practice. We also consider the multi-class classification problem for all 4 classes simultaneously. 

  For binary classification problems, we measure the prediction quality of our algorithm by area under receiver operating characteristic (ROC AUC). This classification quality measure is useful when single number is needed to reflect the performance of the classification algorithm~\cite{Florkowski2008}. The relatively small sample sizes in all the considered problems require the careful choice of evaluation scheme. We use 10-fold cross-validation, where the parameters of machine learning models were tuned based on training data for each test fold, to ensure that we do not use any information from testing data for training. The results were averaged over 100 independent cross-validation partitions.

\subsection{Tools}
  For the numerical experiments, we used Python programming language and Jupyter notebook environment. We did matrix calculations and numerical analysis using NumPy, SciPy, NetworkX and igraph libraries. The main classification pipeline was implemented using scikit-learn library~\cite{Pedregosa2011}, which we used for all the classifiers except GBDT. For Gradient Boosted Decision Trees, we used xgboost library~\cite{Chen2016}. The code is available from authors upon request.

\subsection{Results}
\label{sec: results}
  First, we discuss the results for binary classification problem. We start by reminding the state-of-the-art results for ADNI2 dataset~\cite{Kurmukov2017} obtained by kernel algorithm based on community structure, see Table~\ref{tab: results_sta}.

  \begin{table}
    \centering
    \begin{tabular}{|l|c|c|c|c|}
      \hline
      & AD vs NC & AD vs LMCI & LMCI vs EMCI & EMCI vs NC \\
      \hline
      \hline
      Community (SVM) & $0.831 \pm 0.009$  & $0.762 \pm 0.018$ & $0.523 \pm 0.028$ & $0.628 \pm 0.018$ \\
      \hline
    \end{tabular}
    \caption{Quality of classification for the method based on community structures of networks.}
    \label{tab: results_sta}
  \end{table}

\begin{figure}[h!]
    \centering
    \includegraphics[width=0.7\textwidth]{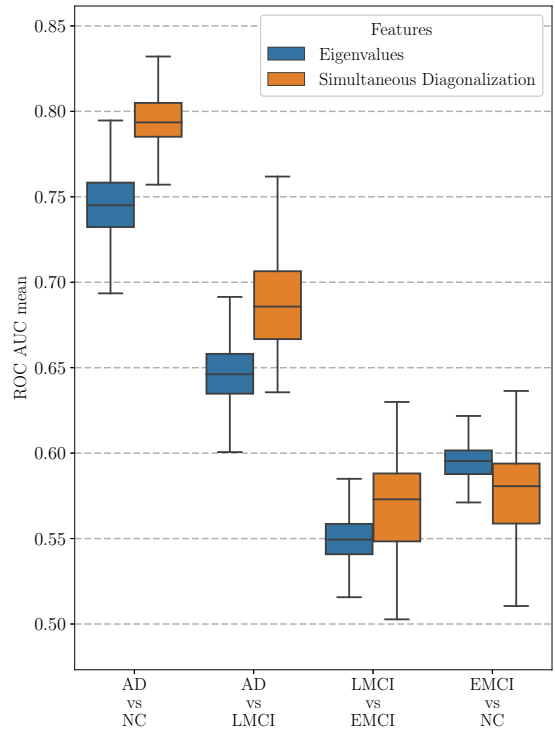}
    \caption{Mean ROC AUC values for best methods for each problem and both feature extraction methods.}
  \label{f1}
  \end{figure}

  Tables~\ref{tab: results_eigen} and~\ref{tab: results_diag} show the mean ROC AUC values for the classification based on eigenvalues \(\adjacencyEigenvalues_i\) and diagonal values of simultaneously diagonalized matrices \(\lambdav_i\) respectively. Figure~\ref{f1} shows the results of best algorithms for each subproblem and for both feature extraction approaches. We observe that for 3 out of 4 classification problems simultaneous diagonalization allows to improve the quality of classification.
 
  \begin{table}
    \centering
    \begin{tabular}{|l|c|c|c|c|}
      \hline
      & AD vs NC & AD vs LMCI & LMCI vs EMCI & EMCI vs NC \\
      \hline
      \hline
      LR & $0.732 \pm 0.017$  & $0.646 \pm 0.018$ & $0.550 \pm 0.014$ & $0.594 \pm 0.011$ \\
      \hline
      SVC & $0.721 \pm 0.023$ & $0.646 \pm 0.016$ & $0.550 \pm 0.015$ & $0.484 \pm 0.047$ \\
      \hline
      SGD & $0.746 \pm 0.021$ & $0.638 \pm 0.017$ & $0.547 \pm 0.017$ & $0.593 \pm 0.017$ \\
      \hline
      RF & $0.667 \pm 0.019$ & $0.586 \pm 0.021$ & $0.449 \pm 0.028$ & $0.489 \pm 0.026$ \\
      \hline
      GBDT & $0.697 \pm 0.020$ & $0.572 \pm 0.029$ & $0.435 \pm 0.023$ & $0.483 \pm 0.029$ \\
      \hline
    \end{tabular}
    \caption{Quality of classification based on eigenvalues for 4 binary classification problems.}
    \label{tab: results_eigen}
  \end{table}

  \begin{table}
    \centering
    \begin{tabular}{|l|c|c|c|c|}
      \hline
      & AD vs NC & AD vs LMCI & LMCI vs EMCI & EMCI vs NC \\
      \hline
      \hline
      LR &  $0.794 \pm 0.017$ & $0.687 \pm 0.026$ & $0.551 \pm 0.014$ & $0.579 \pm 0.024$ \\
      \hline
      SVC & $0.770 \pm 0.018$ & $0.676 \pm 0.019$ & $0.554 \pm 0.033$ & $0.527 \pm 0.037$ \\
      \hline
      SGD & $0.793 \pm 0.018$ & $0.682 \pm 0.024$ & $0.552 \pm 0.038$ & $0.576 \pm 0.026$ \\
      \hline
      RF &  $0.722 \pm 0.022$ & $0.610 \pm 0.033$ & $0.540 \pm 0.030$ & $0.568 \pm 0.017$ \\
      \hline
      GBDT & $0.730 \pm 0.023$ & $0.613 \pm 0.026$ & $0.570 \pm 0.027$ & $0.570 \pm 0.031$ \\
      \hline
    \end{tabular}
    \caption{Quality of classification based on simultaneous diagonalization for 4 binary classification problems.}
    \label{tab: results_diag}
  \end{table}

\section{Conclusions}
\label{sec: conclusions}
  In this work, we show that a simple but powerful approach to simultaneous matrix diagonalization allows us to extract more informative features for classification compared to features extracted from individual adjacency matrices. The study of 4 binary brain disease classification problems shows that proposed approach outperforms baselines based on spectra of graphs and in some cases competes with state-of-the-art approaches to connectome classification. Further work may target the incorporation of a feature extraction method inside the classification algorithm, which may further improve classification quality. Another possibility is to find more advanced approaches for simultaneous feature generation from graphs, for example based on the graph community structure.

\begin{acknowledgement}
  The research was supported by the Russian Science Foundation grant (project 14-50-00150). Some data used in preparing this article were obtained from the Alzheimer's Disease Neuroimaging Initiative (ADNI) database. A complete listing of ADNI investigators and imaging protocols can be found at \url{adni.loni.usc.edu}.
\end{acknowledgement}

\bibliographystyle{spmpsci}
\bibliography{bibl}

\end{document}